\documentclass[manuscript]{acmart}
\pdfoutput=1
\renewcommand\footnotetextcopyrightpermission[1]{} 
\AtBeginDocument{%
  \providecommand\BibTeX{{%
    \normalfont B\kern-0.5em{\scshape i\kern-0.25em b}\kern-0.8em\TeX}}}

\acmConference[FunCausal '23]{}{May 09--12,
  2023}{Grenoble, France}

\acmBooktitle{Proceedings of Fundamental Challenges in Causality, May 09--12, 2023, Grenoble, France}

\usepackage{microtype}
\usepackage{graphicx}
\usepackage{subfigure}
\usepackage{booktabs} 

\usepackage{amsfonts}
\usepackage{amsmath}
\usepackage{xfrac}
\usepackage{enumitem}

\begin{document}

\title{Causal Temporal Graph Convolutional Neural Networks (CTGCN)}
\author{Abigail Langbridge}
\email{abigail.langbridge18@imperial.ac.uk}
\affiliation{%
  \institution{Imperial College London}
  \city{London}
  \country{UK}
  \postcode{SW7 2BX}
}
\author{Fearghal O'Donncha}
\affiliation{%
  \institution{IBM Research Europe}
  \city{Dublin}
  \country{Ireland}
}
\author{Amadou Ba}
\affiliation{%
  \institution{IBM Research Europe}
  \city{Dublin}
  \country{Ireland}
}
\author{Fabio Lorenzi}
\affiliation{%
  \institution{IBM Research Europe}
  \city{Dublin}
  \country{Ireland}
}
\author{Christopher Lohse}
\affiliation{%
  \institution{Trinity College Dublin}
  \city{Dublin}
  \country{Ireland}
}
\author{Joern Ploennigs}
\affiliation{%
  \institution{University of Rostock}
  \city{Rostock}
  \country{Germany}
}

\renewcommand{\shortauthors}{Langbridge et al.}

\setlength{\abovedisplayskip}{6pt}
\setlength{\belowdisplayskip}{6pt}

\begin{abstract}

Many large-scale applications can be elegantly represented using graph structures. Their scalability, however, is often limited by the domain knowledge required to apply them. To address this problem, we propose a novel Causal Temporal Graph Convolutional Neural Network (CTGCN). Our CTGCN architecture is based on a causal discovery mechanism, and is capable of discovering the underlying causal processes. The major advantages of our approach stem from its ability to overcome computational scalability problems with a divide and conquer technique, and from the greater explainability of predictions made using a causal model. We evaluate the scalability of our CTGCN on two datasets to demonstrate that our method is applicable to large scale problems, and show that the integration of causality into the TGCN architecture improves prediction performance up to 40\,\% over typical TGCN approach. Our results are obtained without requiring additional domain knowledge, making our approach adaptable to various domains, specifically when little contextual knowledge is available.
\end{abstract}


\received{10 February 2023}

\maketitle

\section{Introduction}
Numerous real-world processes contain interconnected dynamics characterised by a complex organizational structure. Examples include environmental systems, energy grids, and epidemic outbreaks. Many classical mathematical frameworks exist to model these dynamics such as Navier-Stokes or convective-diffusion equations. While these explicitly encode known relationships, machine learning provides opportunity to resolve established and latent dynamics.

Graphs have been historically used to model many of the underlying structures and physical behaviours of spatial systems such as road networks \cite{thomson1995graph}, building thermodynamics \cite{Tabunschikov92}, and water and energy grids \cite{mackaness1993use}. This allows practitioners to explicitly encode their domain knowledge. A fundamental assumption in these models is that the underlying structure of these dynamics are well known to be modelled. But, most real-world applications do not admit a-priori knowledge of these structures. Therefore, we aim to create a graph model that learns this structure using causal discovery.

Recent Temporal Graph Convolutional Neural Networks (TGCN) \cite{sanchez2018graph,dai2021graph} utilise this available domain knowledge in graphs by combining learning over temporal and graphical features.
These TGCN models also assume that spatial information or similar prior knowledge regarding connections is available \cite{pvnetwork_spatial, Yu2018SpatioTemporalGC}.
However, in practise, many use cases for spatiotemporal modelling do not have well-defined graph structures. In contrast, in most practical applications we see clients collect new timeseries data faster than they are able to specify contextual information which grows the problem.

This lack or uncertaincy of a-priori spatial knowledge is intensified for downstream tasks of the TGCN. First, incorrect graph models will strongly influence the prediction performance of the model limiting optimization and diagnostic tasks. Second, recent works have investigated the efficacy of post-hoc explanations for graph convolutional neural network (GCNN) predictions \cite{GNNExplainer}, with some approaches using causal inference methods \cite{OrphicX}, but these fundamentally rely on the correctness and completeness of the graph over which they are explaining.

Some work has investigated treating the graph as a learnable parameter to optimise for a given downstream task \cite{dgm_learning, latent_graph_learning}. \citet{adaptive_dep_lrn} proposed an efficient method to construct a dynamic dependency graph based on statistical structure learning models. Prominent limitations of these methods is that they are prone to learning spurious correlations and not the underlying causes. Other work looks into identifying the graph from physics equations from scientific papers utilizing NLP approaches \cite{ba2022automated}, but it assumes that systems exhibit static physical behaviour.

Also traditional Bayesian structure learning approaches are often not applicable to high-dimensional, real-world data due to their computational cost and unrealistic assumptions which include stationarity, the absence of latent confounders in data, and that none of the underlying relationships are contemporaneous \cite{bayesian_cost}.

In this work, we present a novel, scalable method for deducing causal relationships in large observational data, which we integrate into a TGCN architecture to overcome the limitation of requiring a-priori domain knowledge. This gives our causal-informed TGCN (CTGCN) visibility of the underlying structural causal model in an automated way and facilitates its self-adaptation and scalability in diverse large-scale applications. Furthermore, we investigate the computational complexity and predictive skill of the causal discovery process in order to improve the performance for large scale, real-world applications and non-IID datasets. We further study the effects of decomposing the causal discovery problem on different downstream predictive task, as we posit that integrating structural causal models into the graph convolution layer of a TGCN model improves model performance, robustness and explainability.

The main contributions of this work are:
\begin{itemize}[noitemsep,topsep=0pt]
    \item We develop a novel causal-informed TGCN discovery method in order to facilitate adoption for large-scale, interconnected, and complex applications.
    \item We extend existing causal discovery methods with spatial and temporal decomposition to improve their scalability for large-scale applications.
    \item We demonstrate that the integration of causal structure information into predictive models through a graph convolution layer improves forecasting performance.
\end{itemize}

\section{Related Work}
First, we provide a background on spatiotemporal graph convolutional neural networks (GCNN). We then discuss the utility of existing causal inference methods for large-scale data.

\subsection{Spatiotemporal GCNN}

Our approach relies on GCNN \cite{KipfW17}, initially introduced by \citet{Bruna2014SpectralNA}, and extended by \citet{Duvenaud2015ConvolutionalNO} with fast localised convolutions.
Approaches such as TGCN \cite{Yu2018SpatioTemporalGC,li2016gated,8809901} augment this method with sequence-to-sequence learning to encode temporal dynamics.
The penetration of GCNN into sensor-driven applications using time series data led to the extension of GCNN with sequence-to-sequence learning methods such as Gated Recurrent Unit (GRU) or Long Short-Term Memory (LSTM) for forecasting applications \cite{pathak2018forecasting}. The combination of GCNN with sequence-to-sequence learning methods is generally motivated by the need to simultaneously capture spatial and temporal dependencies.
In this case, the GCNN is used to capture spatial dependencies by building a filter that acts on the nodes and their $n$-th order neighbourhood (usually first order). This enables the filter to capture spatial features between the nodes, and further the GCNN can be developed by stacking multiple convolutional layers. The sequence-to-sequence learning method is employed to capture dynamic changes of the systems.

However, modelling complex topological structures with sequence-to-sequence learning usually fails in capturing a system's underlying causal processes and thus restricts the input-output relationships to correlations. Schölkopf highlights the robustness of similar approaches as a key problem, suggesting the integration of causal modelling in ML could overcome this \cite{causal4ml}. Many real-world problems that we are interested in studying violate the IID assumption, which underpins many correlational approaches. 
We propose an extension of the TGCN architecture that introduces a scalable causal convolution to capture the characteristics of the underlying system.


\subsection{Causal Discovery}

Time-series data has long been a challenging problem in causal discovery: while the canonical order of data facilitates the directing of causal links (thus overcoming Markov equivalence), strong autocorrelation and the presence of unobserved confounders reduces detection power~\cite{PCMCI+}. The scale of most modern data exacerbates these problems, rendering many methods intractable.


Conditional independence (CI) methods have shown significant promise in time-series causal discovery due in part to their flexibility to utilise various CI tests. 
The PC algorithm presented in the seminal work \citet{pc} utilises sparsity to improve the scalability of the CI method. This has led to incremental improvements on the base PC algorithm, firstly in the form of the Fast Causal Inference (FCI) algorithm and its adaptation to time series data \cite{fci,ts_fci}, and further for improving detection power with PCMCI$^+$ \cite{PCMCI+} and latent variable detection with LPCMCI \cite{LPCMCI}.


These approaches work well for small scale problems, but face serious combinatorial explosion problems which renders these promising approaches infeasible for large scale applications. 
In order for our proposed causal TGCN to be tractable for real-world systems, a more scalable approach to causal discovery must be developed. The score-based fast Greedy Equivalence Search (fGES) \cite{FGES} highlights the appetite for scalable causal discovery algorithms, however to the best of the authors' knowledge no equivalent approach to the acceleration of CI-based methods exists, and score-based methods are limited by their inability to overcome Markov equivalence.


\section{Methodology}

To overcome the challenges in applying TGCN at large scale, we developed a novel causality-based TGCN approach that is capable of automatically identifying the causal relationships of spatiotemporal systems and configuring TGCN models accordingly.

\begin{figure*}[t!]
\vskip 0.2in
\begin{center}
\centerline{\includegraphics[width=0.9\textwidth]{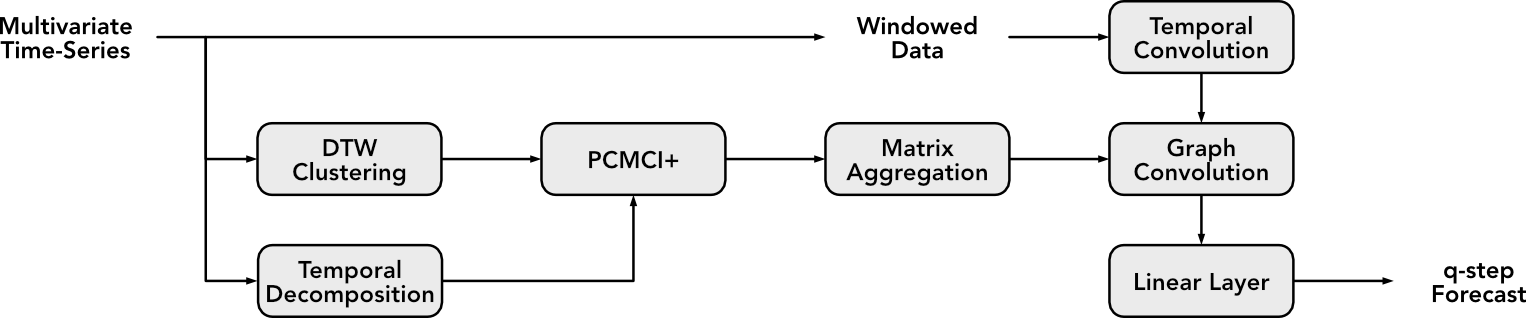}}
\caption{Block diagram of the proposed causal TGCN architecture.}
\label{fig:block-diagram}
\end{center}
\vskip -0.2in
\end{figure*}

Figure~\ref{fig:block-diagram} shows the workflow of the methodology. We use as input a multivariate timeseries dataset, and first do causal discovery using PCMCI$^+$ \cite{PCMCI+}. To overcome the underlying scalability problems, we decompose the identification problem temporally and spatially and then recombine the results in a matrix aggregation step. Then we transform the discovered causal relationships into the adjacency matrix of a TGCN.

\subsection{Graph Convolutional Neural Networks}\label{sec:gcnn}


We consider as input a feature matrix $\mathcal{X} \in \mathbb{R}^{N \times P}$ of a multi-variate spatiotemporal system, with the number of features $N$, the length of the time series $P$ and an observation vector $\mathcal{X}_{t} \in \mathbb{R}^{N}$ at time $t$.

Each feature $\mathcal{X}^j \in \mathcal{X}$ can be represented as node in a graph $\mathcal{G}$. As our first modification of a traditional GCNN model, we define a \textit{causal adjacency matrix} $A$ that defines the causal relationship for each directed edge $a_{j,k}$ between feature pairs $\mathcal{X}^j$ and $\mathcal{X}^k$. We distinguish two forms: (i) first a binary causal adjacency matrix where the relationship $a_{j,k} \in \{0,1\}$ defines the binary existence of a causal link. We also consider (ii) a weighted causal adjacency matrix, where $a_{j,k} \in \mathbb{R}^{+}_0$ provides the weight of the causal relationship assuming that any $a_{j,k}=0$ signifies no causal relationship.

To evaluate the performance of our approach, we define the spatiotemporal forecasting problem as learning the mapping function $\operatorname{f}$ using the structural information provided by the adjacency matrix $A$ and the features $\mathcal{X}$.

A GCNN model constructs the mapping function $\operatorname{f}$ as filter in the Fourier domain. The filter then acts on the nodes of the graph and its first order neighbourhood. This allows the topological structure and the spatial features between the nodes to be captured. Subsequently, the GCNN model can be established by stacking multiple convolutional layers. The GCNN is given by
\begin{align}
\operatorname{f}\left(\mathcal{X}, A\right) = \sigma\left(\hat{A}\, \operatorname{ReLU}\!\!\left(\hat{A}\,\mathcal{X} W^{(0)}\right)\,W^{(1)}\right),
\end{align}

where $\mathcal{X}$ is the feature matrix, $A$ represents the adjacency matrix, $\hat{A} = \tilde{D}^{-\frac{1}{2}}\tilde{A} \tilde{D}^{-\frac{1}{2}}$ is a preprocessing step, $\tilde{A} = A+ I_{N}$ is a matrix that considers the features of the nodes for which the learning is conducted, $\tilde{D}$ is a degree matrix, where $\tilde{D} = \sum_{j} \tilde{A}_{ij}$, $W^{(0)}$ and $W^{(1)}$ represent the weight matrix in the first and second neighborhood, and $\sigma$, and $\operatorname{ReLU}$ represent the activation functions.

\subsection{Causal Inference of the Adjacency Matrix}

To solve the above GCNN problem, we need to identify the adjacency matrix $A$. We frame this as a causal learning problem on the structure of the underlying causal processes:
\begin{equation}
    \mathcal{X}^j_t = \operatorname{g}_j\left(\mathcal{P}(\mathcal{X}^j_t), \eta^j_t \right),
\label{eq:scm}
\end{equation}
where $\operatorname{g}_j$ are arbitrary measurable functions that depend non-trivially on the causal parents $\mathcal{P}$ of the given node $j$, and $\eta$ is independent noise obscuring the causal processes.

As described in Section 1.3, there are various methods for discovering these causal relationships. In this work, we adapt the constraint-based method PCMCI$^+$ presented by \citet{PCMCI+}. This method increases the detection power over highly autocorrelated data compared to seminal methods \cite{pc}, and its ability to detect contemporaneous links makes it particularly suitable for dealing with noisy real-world spatiotemporal data.


Causal discovery methods such as PCMCI$^+$ are bound by assumptions that the observations are faithful to and fully representative of the underlying processes, and that these processes are stationary and acyclic. PCMCI$^+$ is based on Runge's definition of a momentary conditional independence (MCI) test \cite{PCMCI} that, for each lagged observation $\mathcal{X}^j_{t-\tau}$, $\mathcal{X}^k_t$ of a feature pair $j,k$, tests for the existence of a causal relationship given a lag $\tau \in (1,\dots,\tau_{max})$. If the $p$-value of the test is above the significance threshold $\alpha$, we consider a binary causal relationship between $\mathcal{X}^j_{t-\tau}$, $\mathcal{X}^k_t$ where $c_{j,k,t,\tau}=\operatorname{L}(p_{j,k,t,\tau}>\alpha)$ with the logical function $L(b)$ which is $1$ if condition $b$ evaluates true and $0$ otherwise.

The biggest limitation of the approach is its scalability. As an MCI test is computed for each lagged combination of features, we observe a worst-case time complexity of $\mathcal{O}(P \cdot ((N \cdot \tau_{max})^2 + e^N))$ for PCMCI$^+$. While this is a significant improvement on the original PC algorithm's worst case $\mathcal{O}(P \cdot e^{N \cdot \tau_{max}})$, it is still intractable for large $N$ and $P$. The choice of CI tests also affects the runtime, with methods more robust to non-linear relationships, and therefore more generalisable, increasing computational cost.

\subsection{Temporal Decomposition}
A key limitation of existing causal discovery methods is the assumption of stationarity, which is unrealistic for data spanning months or years and encompassing varying temporal dynamics. To overcome this, and improve the scalability of our method, we propose splitting the data along the time axis into approximately stationary periods with $P_T$ observations.

These periods are typically derived based on generalised domain expertise or statistical analysis of exogenous features. Human systems such as buildings, energy, transport, or finance often exhibit daily, weekly, or monthly patterns, while natural systems such as oceans, agriculture, and weather often exhibit daily, seasonal, or annual dynamics. 


\subsection{Spatial Decomposition}



Modern systems in manufacturing, smart cities, and the automotive industry are highly monitored with thousands of IoT devices. The large spatial dimensionality is computationally challenging for existing CI algorithms. Intelligent spatial decompostion can dramatically reduce computational expense. Many decomposition approaches exist such as statistical, domain-inspired, or rule-based methods. 

Dynamic time warping (DTW) is a pattern-matching approach to the alignment of time-series data first proposed for speech recognition by \citet{DTW}. It has more recently been popularised as a method for the unsupervised classification of large temporal data in applications from finance to astronomy \cite{cluster-review}. The principal advantage of the DTW approach over Euclidean distances lies in the fact that one-to-many relationships can be mapped between candidate timeseries, which facilitates the identification of shared trends even if they occur on different timescales.

By utilising unsupervised DTW clustering on the feature axis, we can decompose the causal discovery problem into several sub-problems which we solve independently. We evaluate PCMCI$^+$ within each cluster and for each temporal period. This divide-and-conquer approach reduces the worst-case complexity of PCMCI$^+$ to $\mathcal{O}(D \cdot P_T \cdot (( N_C \cdot \tau_{max})^2 + e^{N_C}))$ with $P_T$ being the temporal decomposition period, $D$ being the cluster number and $N_C$ being the maximum cluster size.
We posit that by clustering in this way, we minimise the number of cross-cluster relationships in the underlying causal model, and therefore maximise the recall of the decomposed problem.

Further, as DTW clustering is an unsupervised method, this decomposition requires few additional parameters to be run automatically as shown in Table \ref{tab:ts-pcmci}.



\subsection{Adjacency Matrix Construction} \label{sec:adj}

The causal discovery steps outlined above produce results $c_{j,k,t,\tau}=\operatorname{L}(p_{j,k,t,\tau}>\alpha)$ for each $\tau$-lagged timestep $t$ and detect causal relationships which span from contemporaneous ($\tau = 0$) up to a maximum lag ($\tau_{max}$) which exceed some significance threshold $\alpha$. These parameters and their selection are summarised in Table \ref{tab:pcmci-param}.

After the temporal and spatial decomposition we retrieve $c_{j,k,t,\tau}$ for all temporal and spatial clusters and aggregate them in form of our causal adjacency matrix $A$. We first aggregate all test results within each temporal $\mathcal{T}$ and spatial $\mathcal{S}$ sample set and perform a majority vote
\begin{equation}
\hat{c}_{j,k,\mathcal{T},\mathcal{S}} = \operatorname{MV}\!\!\left(\sum_{t=0}^{P_T} \sum_{\tau=0}^{\operatorname{min}(t,\tau_{max})}\!\!\!\!\!\!\!\!\frac{1}{P_T (\tau_{max}-1)} c_{j,k,t,\tau}\!\!\right)\!\!,
\label{eq:agg}
\end{equation}
with the majority voting function $\,\operatorname{MV}(v)$ that is $1$ if $v>0.5$ and $0$ otherwise.
This filters out causal relationships that were only discovered in specific time steps, but, are not common within a temporal and spatial sample set. We then aggregate the votes from all temporal and spatial sample sets in our causal adjacency matrix $A$. We use the following aggregation strategies: 
\begin{equation}
\begin{aligned}
a_{j,k}^\text{ANY;W} = \sum_{\mathcal{S}} \sum_{\mathcal{T}} \hat{c}_{j,k,\mathcal{T,S}},\\
a_{j,k}^\text{MT;W} = \operatorname{MV}(T^{-1} a_{j,k}^\text{ANY;W})\cdot a_{j,k}^\text{ANY;W},\\
a_{j,k}^\text{ANY;UW} = \operatorname{B}(a_{j,k}^\text{ANY;W}),\\
a_{j,k}^\text{MT;UW} = \operatorname{B}(a_{j,k}^\text{MT;W}).
\end{aligned}
\label{eq:agg2}
\end{equation}
The weighted aggregation $a_{j,k}^\text{ANY;W}$ computes the total sum of causality test results. The weighted majority vote $a_{j,k}^\text{MT;W}$ considers the weight of relationships occurring in a majority of sample sets in the number $T$ of temporal sets. The unweighted aggregations $a_{j,k}^\text{ANY;UW}$ and $a_{j,k}^\text{MT;UW}$ reduce the weighted adjacency matrix to a binary one using the binary function $\,\operatorname{B}(v)$ that is $1$ if $v>0$ and $0$ otherwise.
If we do not model the directed behaviour of the system, we can simplify this to an undirected matrix via $a_{j,k}=\operatorname{B}(a_{j,k}+a_{k,j})$.



\subsection{Evaluating Performance}\label{sec:prediction}
Forecasting is a common task in spatiotemporal applications for monitoring, anomaly detection, optimisation, etc. 
We consider the task of windowed forecasting, where a model is trained to infer the subsequent $q$ measurements 
given a finite history of length $\lambda$.

Despite the growing popularity of sequence-to-sequence learners for forecasting applications, we consider the foundational case of a simple one-dimensional convolution along the time axis to capture the temporal behaviour of the system. This has the dual benefit of reducing training overhead for improved scalability and clearly demonstrating the effectiveness of our method even with a simplistic architecture. The simpler network structure also enables explainability \cite{GNNExplainer} that is important for practical applications.

We design a forecasting architecture which begins with a temporal convolution, followed by the causal graph convolution layer. We posit that this ordering allows the temporal dynamics of the system to be captured at each node, and then these distilled features can be used to inform forecasting at causally-related nodes.


The TGCN we employ builds on \citet{KipfW17} and is implemented in PyTorch Geometric \cite{pyg}. A final linear layer produces a forecast of length $q$. 
The model is trained in batches using RMSE loss. Hyperparameters are tuned for each dataset using the adaptive grid search method provided by Optuna \cite{optuna}.

\section{Experiments}
We demonstrate the performance of our method in two contexts, selected to demonstrate the generalisability of the approach.
Experiments were conducted on an Apple M1 Max 32GB MacBook Pro (2021) running MacOS Monterey 12.6 and Python 3.9.12.

Code to replicate our experiments is provided at \url{https://tinyurl.com/ctgcn-code} and will be open-sourced

\subsection{Building Heating System}
The first dataset is heterogeneous and multivariate, and generated from a simulated heating, ventilation, air conditioning (HVAC) system which controls the temperature of two rooms and an interconnecting corridor \cite{ploennigs2017semantic}. It consists of data from thirty sensors measuring the system variables including the internal status of the system's boiler, chiller and ventilation units, the temperature within each room and externally, and room occupancy. The data is sampled every 10 minutes over a total period of one year.

We included this dataset in our experiments as, due to its simulated nature, we have a ground truth about the causal relationships which we can use to investigate the quality of the causal discovery, as well as to measure the performance of the CTGCN on the downstream forecasting task. The dataset is further strongly heterogenous with datapoints representing sensors with different characteristics and value ranges. It is known that this is a challenging task for GCNNs \cite{zhao2021heterogeneous,ba2022automated}, and may also adversely effect the causal discovery, rendering this a particularly interesting study.

The ground truth adjacency matrix is defined based on the underlying physics of the simulation \cite{wetter2014modelica}, and contains 50 relationships between the 30 sensors.

For this dataset we evaluate performance for both temporal and spatiotemporal decomposition in terms of the accuracy of causal discovery, the performance of forecasting the temperature of one of the rooms, and the runtime.

Our temporal decomposition consists of a daily temporal split with $P_T = 144$ based on auto-correlation analysis, yielding 365 causal sub-problems. We select one hour ($\tau_{max}=6$) as the maximal time lag based on system dynamics. For the spatial decomposition we use 10 clusters after running an elbow test.

\subsection{Highway Traffic Flow}
We further verify our method on homogeneous real-world traffic flow data collected by the California Department of Transport. The data consists of 30-second traffic flow speed measurements collected from 228 locations across the California state highway system over weekdays in May and June 2012. We downsample the data to five-minute intervals to align with work by \citet{Yu2018SpatioTemporalGC} and facilitate comparison. In that work, Yu et al. also proposes a distance-thresholding mechanism to identify the adjacency matrix
\begin{equation}
w_{ij} =
\begin{cases}
    exp\!\!\left(-\frac{d^2_{i,j}}{\sigma^2}\right)\!,&\!\!\text{if } i \neq j \text{ and } exp\!\!\left(-\frac{d^2_{i,j}}{\sigma^2}\right)\!\geq\!\epsilon;\\
    0,              &\!\!\text{otherwise}.
\end{cases}
\label{eq:distance-adj}
\end{equation}
We evaluate the effectiveness of our method against their results using their parameters $\sigma^2 = 10$ and $\epsilon = 0.5$.

To overcome the non-stationarity of the data, we split the data by date, conducting causal discovery for each of the 44 days of data. We also conduct spatial decomposition with 25 clusters estimated using an elbow test. The maximal lag we consider is $\tau_{max} = 9$ to align with \citet{Yu2018SpatioTemporalGC}.

For this dataset, we demonstrate the runtime improvement of spatiotemporal decomposition over temporal, and evaluate the compromise between runtime and prediction accuracy.

\section{Results}
\subsection{Building Heating System}
We first compare the ground-truth with the temporal TGCN and the spatiotemporal (DTW clustered) TGCN approach with different aggregation strategies. Table~\ref{tab:results-building} shows the precision, accuracy and F1 score for the different approaches outlined in Section \ref{sec:adj}.

The accuracy of all approaches is higher than 83\,\% due to a high true negative rate as the association matrix is sparse, hence precision is a more relevant metric. Precision is only 14.2\,\% to 16.6\,\% for the temporal approach, which may be surprising as this approach should be able to discover all potential causal relationships. However, this results in a high false positive rate with about 150 causal relationships discovered. The spatiotemporal approach has a significantly higher precision with 32.8\,\% to 39.2\,\% discovering an average of 25 relationships. Despite this, the method is not able to discover all relationships due to the heterogenous nature of the dataset and our spatial decomposition. We see that data are grouped semantically by the DTW clustering step, e.\,g.\ separate clusters are created for temperature, CO2, occupancy and humidity. As our algorithm does not consider causal relationships to exist between clusters, we miss these relationships (See Appendix~\ref{sec:graph} for details). Nonetheless, the spatiotemporal approach has a higher accuracy, precision and a significantly lower compute time. The temporal approach took 288 hours (12 days) to compute, while the spatiotemporal approach computed in 49 hours (2 days, see Table~\ref{tab:runtimes}).

The matrix aggregation step also improves the result. One performance improvement approach could be to sample individual days from the dataset to estimate the adjacency matrix. This approach (AVG) has a mean precision of 32.8\,\%. But, analysing the full dataset and applying the Majority Threshold (MT) or the ANY aggregation we improve precision to 38.4\,\% and 39.2\,\%, respectively.

\begin{table}[t]
\caption{Results for the building dataset.}
\label{tab:results-building}
\begin{center}
\begin{small}
\begin{sc}
\vskip -0.15in
\begin{tabular}{{p{0.41\columnwidth} p{0.14\columnwidth}  p{0.14\columnwidth} p{0.10\columnwidth}}} 
\toprule
Approach          & Precision & Accuracy & F1\\
\midrule
Temporal AVG   & 14.2\,\%  & 83.1\,\%   & 14.1\,\% \\
Temporal  MT   & 16.6\,\%  & 84.7\,\%   & 15.2\,\% \\
Temporal ANY   & 15.2\,\%  & 83.9\,\%   & 14.6\,\% \\
\midrule
Spatiotemporal   AVG   & 32.8\,\%  & 88.2\,\%   & 21.9\,\% \\
Spatiotemporal   MT    & 38.4\,\%  & 88.8\,\%   & 26.3\,\% \\
Spatiotemporal   ANY   & 39.2\,\%  & 88.8\,\%   & 28.2\,\% \\
\bottomrule
\end{tabular}
\end{sc}
\end{small}
\end{center}
\vskip -0.2in
\end{table}

We further compare the prediction performance of the discovered adjacency matrix to the ground truth when forecasting the room temperature. This has multiple causal relationships and relevant practical applications in energy management. As detailed in Section \ref{sec:prediction}, we utilise a CTGCN that is configured with the different association matrices resulting from the temporal and spatiotemporal clustering and the different aggregation methods for prediction. Table~\ref{tab:pred-results-building} compares the results. We provide also the RMSE for an unconnected and fully connected TGCN for context.

The RMSE for the ground truth prediction is significantly better than the unconnected and fully connected TGCN demonstrating the well-known benefits of using domain knowledge in TGCN configuration.
The unweighted temporal and spatiotemporal CTGCN outperform the unconnected TGCN but not the ground truth. This is expected given that not all causal relationships were discovered. However, the temporal and spatiotemporal CTGCN is able to outperform the ground truth TGCN when weighted by frequency. The frequency clearly encodes important information about the relevance of a causal relationship that is not contained in the binary ground truth, and these weights allow the TGCN to compensate for the missing edges. In result, the spatiotemporal weighted majority threshold had the lowest RMSE, showing that our CTGCN can outperform the ground truth in a prediction problem.

\begin{table}[t]
\caption{Comparison of the prediction accuracy of different aggregation methods for the building dataset.}
\label{tab:pred-results-building}
\begin{center}
\begin{small}
\begin{sc}
\vskip -0.15in
\begin{tabular}{lccc}
\toprule
Approach & RMSE\\
\midrule
Ground Truth TGCN                      & \textbf{0.8297}\\
\midrule
Unconnected TGCN                      & 0.8981\\
Fully connected TGCN                    & 0.8548\\
\midrule
Temporal MT Unweighted CTGCN             & 0.8668\\
Temporal MT Weighted CTGCN             & 0.8469\\
Temporal ANY Unweighted CTGCN            & 0.8735\\
Temporal ANY Weighted CTGCN              & \textbf{0.8250}\\
\midrule
Spatiotemporal MT Unweighted CTGCN  & 0.8668\\
Spatiotemporal MT Weighted CTGCN   & \textbf{0.8209}\\
Spatiotemporal ANY Unweighted CTGCN & 0.8306\\
Spatiotemporal ANY Weighted CTGCN  & 0.8735\\
\bottomrule
\end{tabular}
\end{sc}
\end{small}
\end{center}
\vskip -0.2in
\end{table}

\subsection{Highway Traffic Flow}


The mean runtime of each day of the temporally-decomposed traffic data was 35 hours, corresponding to an estimated total runtime of more than 64 days (Table \ref{tab:runtimes}). As such, causal discovery was not conducted for every temporal split with this method. Instead, we selected the first, last, and central five days of data to get an overview of the entire period and include variations in causal relationships over time. Forecasting performance was measured across the entire dataset to test the representativity of the causal discovery method when forecasting on unseen data.



\begin{figure}[t]
\begin{center}
\centerline{\includegraphics[width=0.9\columnwidth]{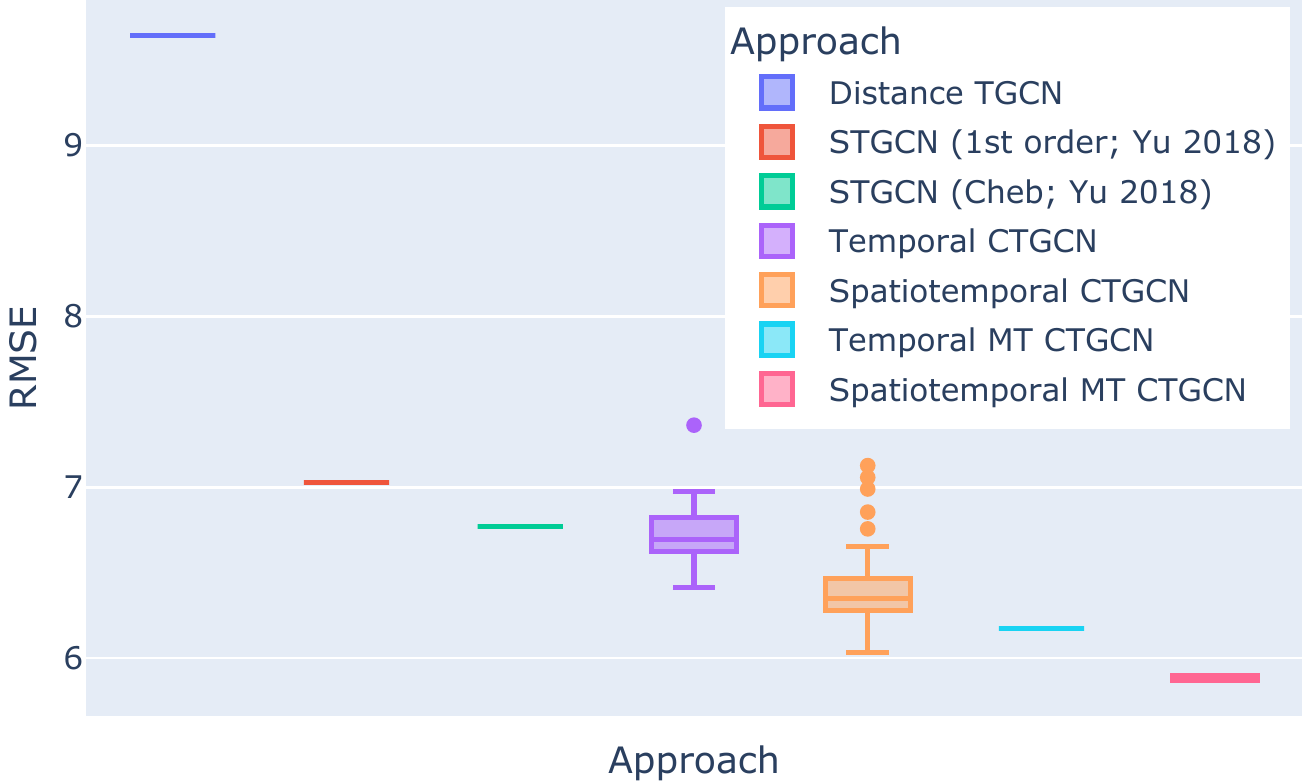}}
\vskip -0.15in
\caption{Forecasting RMSE on the traffic dataset compared to STGCN \cite{Yu2018SpatioTemporalGC}}
\label{fig:decomp-performance}
\end{center}
\vskip -0.3in
\end{figure}

As evidenced in Fig.~\ref{fig:decomp-performance} and Appendix Table \ref{tab:pred-results-traffic}, downstream performance is significantly improved by introducing the temporally decomposed causal adjacency matrix into the TGCN architecture. Forecasting RMSE was on average 30.4\% improved over the benchmark, with the best-performing and worst-performing days yielding 33.5\% and 23.6\% improvements respectively.



To combine the causal graphs calculated from different days of data, the majority threshold aggregation method was used. This aggregation improves the performance of the temporal decomposition, resulting in an RMSE 36.0\,\% lower than the benchmark, and lower than any of the daily graphs. This result is notable as it demonstrates that by combining causal graphs from different sections of the data we can better capture the overall causal relationships.

To compare our results to a state-of-the art approach, we added the STGCN results from \citet{Yu2018SpatioTemporalGC}. Note that the RMSE of the same distance based benchmark is higher than the STGCN due to our simpler sequence-to-sequence learner. But, the improved adjacency matrix of the Temporal MT graph outperforms even the more advanced STGCN.


\begin{table}[t]
\caption{Comparison of causal discovery runtimes in hours for the datasets.}
\label{tab:runtimes}
\begin{center}
\begin{small}
\begin{sc}
\vskip -0.15in
\begin{tabular}{p{0.2\columnwidth} p{0.2\columnwidth} p{0.23\columnwidth} p{0.15\columnwidth}}
\toprule
Dataset            & Temporal  & Spatiotemporal & Factor\\
\midrule
Building           & 287.7\,h  & 49.8\,h & 5.8\,x \\
Traffic       & 1540.0\,h  & 106.2\,h & 14.5\,x \\
\bottomrule
\end{tabular}
\end{sc}
\end{small}
\end{center}
\vskip -0.25in
\end{table}


As shown in Table \ref{tab:runtimes}, introducing spatial decomposition accelerates causal discovery on this data by more than 14 times. 
Despite this, Fig.~\ref{fig:decomp-performance} demonstrates that predictive performance was not negatively affected, with performance improving in some cases. The minimum, mean and maximum improvement over the benchmark are 26.1\%, 33.5\% and 37.4\% respectively.
Spatiotemporal performance is further improved through MT aggregation, with RMSE 39.1\% lower than the benchmark and 4.9\% lower than Temporal MT.





\section{Discussion}

Our experiments have shown that using an automatically discovered causal adjacency matrix with a TGCN architecture can improve the forecasting performance over typical spatial approaches. Further, we demonstrate that through context-aware decomposition of the causal discovery problem we accelerate the discovery process such that previously intractable problems can be computed on a single machine. This is particularly notable for the traffic dataset, which is accelerated 14 times when the problem is both spatially and temporally decomposed over a temporal decomposition.

Importantly, our approach reports high predictive skill with a simplistic sequence-to-sequence learner. Due to the lightweight architecture---constructed of a simple 1-D convolution and a graph convolution layer---we posit that the predictions from this model are more interpretable than more complex SOTA architectures \cite{zeng2022transformers}. Simple weight visualisation or gradient based approaches can provide efficient insight into model performance for such 1-layer systems \cite{nguyen2019understanding}.  
A notable advantage of the proposed framework is that a causal model accurately representing underlying system dynamics allows us to retain performance with the lightweight architecture.

Causal discovery can generate a more practical representation of  latent and contemporaneous relationships. The power of GNNs has been demonstrated across many applications in recent years such as estimating travel times in Google Maps, powering content recommendations in Pinterest, and providing product recommendations in Amazon \cite{velivckovic2023everything}. We propose an agnostic and scalable framework for effective graph generation for such applications. 

We also illustrate that the method used to aggregate causal results from sub-problems is important, with the weighted majority threshold method yielding performance improvements over any of its constituent graphs. We posit that this is due to its ability to to capture information about the relevance of most common causal relationships in the data and filters non-stationary relationships.

However, our results are not exhaustive: we suggest that by implementing a more balanced clustering method, we might see improved precision. Also, the scalability remains a problem for very large datasets ($N >> 1000$) with dense causal models where the spatial decomposition capability is limited. There also remains opportunity to expand this work toward the explainability of the CTGCN model.


\section{Conclusion} 
While TGCN demonstrates excellent forecasting skill across a wide variety of applications, the need for manually configured adjacency matrices hinder large-scale deployments. We demonstrate that causal discovery methods generate a more robust graph structure that better capture system dynamics, and demonstrate improved predictive skill on two real-world datasets and in comparison to state of the art.
We address computational scalability and non-stationarity by implementing efficient spatial and temporal decomposition.

We show that CTGCN demonstrates particularly impressive performance when data is decomposed based on stationarity, causal relationships are calculated for each sub-problem, and aggregated to incorporate these temporal dynamics.
Future work in this space is in the direction of improving the clustering method, automatic parameter identification, and an evaluation of explainability.

\newpage
\pagenumbering{roman}
\renewcommand\thefigure{\thesection\arabic{figure}} 
\renewcommand\thetable{\thesection\arabic{table}} 
\setcounter{figure}{0}
\setcounter{table}{0} 

\appendix

\section{Causal TGCN Parameters}

Tables \ref{tab:pcmci-param} to \ref{tab:ts-pcmci} detail the parameters required to decompose the causal discovery problem and give a brief summary of the selection method for each parameter.
\begin{table}[h!]
\caption{Parameters necessary for the PCMCI$^+$ algorithm.}
\label{tab:pcmci-param}
\begin{center}
\begin{small}
\begin{sc}
\begin{tabular}{{p{0.2\columnwidth} p{0.65\columnwidth}}}
\toprule
Parameter & Selection Method\\
\midrule
$\tau_{max}$ & Domain knowledge. Method is robust to overestimates at cost of long runtimes.\\
\midrule
$\alpha$ & Significance threshold for causal link detection.\\
\midrule
CI test & Contextual knowledge of the likely nature of causal relationships.\\
\bottomrule
\end{tabular}
\end{sc}
\end{small}
\end{center}
\vskip -0.2in
\end{table}

\begin{table}[h!]
\caption{Additional Parameters necessary for the time-decomposed modification of PCMCI$^+$.}
\label{tab:t-pcmci}
\begin{center}
\begin{small}
\begin{sc}
\begin{tabular}{{p{0.2\columnwidth} p{0.65\columnwidth}}}
\toprule
Parameter & Selection Method\\
\midrule
Period & Data analysis \& context.\\
\midrule
Aggregation method & Downstream performance.\\
\bottomrule
\end{tabular}
\end{sc}
\end{small}
\end{center}
\vskip -0.2in
\end{table}

\begin{table}[h!]
\caption{Additional parameters necessary for the space- decomposed modification of PCMCI$^+$.}
\label{tab:ts-pcmci}
\begin{center}
\begin{small}
\begin{sc}
\begin{tabular}{{p{0.2\columnwidth} p{0.65\columnwidth}}}
\toprule
Parameter & Selection Method\\
\midrule
Number of clusters & Elbow plot.\\
\bottomrule
\end{tabular}
\end{sc}
\end{small}
\end{center}
\vskip -0.2in
\end{table}

\section{Traffic Flow Results}
Numeric results for the traffic flow case study are given in Table \ref{tab:pred-results-traffic}.

\begin{table}[ht]
\caption{Comparison of the prediction accuracy of different aggregation methods for the traffic flow dataset.}
\label{tab:pred-results-traffic}
\begin{center}
\begin{small}
\begin{sc}
\begin{tabular}{lccc}
\toprule
Approach & RMSE\\
\midrule
Distance Benchmark   TGCN               & 9.64\\
\midrule
STGCN (1st order; \citet{Yu2018SpatioTemporalGC}) & 7.03\\
STGCN (Cheb; \citet{Yu2018SpatioTemporalGC})      & 6.77\\
\midrule
Temporal CTGCN & 6.73 \\
Spatiotemporal CTGCN & 6.41 \\
\midrule
Temporal MT CTGCN & 6.17 \\
Spatiotemporal MT CTGCN& \textbf{5.88} \\
\bottomrule
\end{tabular}
\end{sc}
\end{small}
\end{center}
\end{table}



\section{Causal Graph for the Building Dataset}
\label{sec:graph}

\begin{figure*}[hb]
\begin{center}
\centerline{\includegraphics[width=\textwidth]{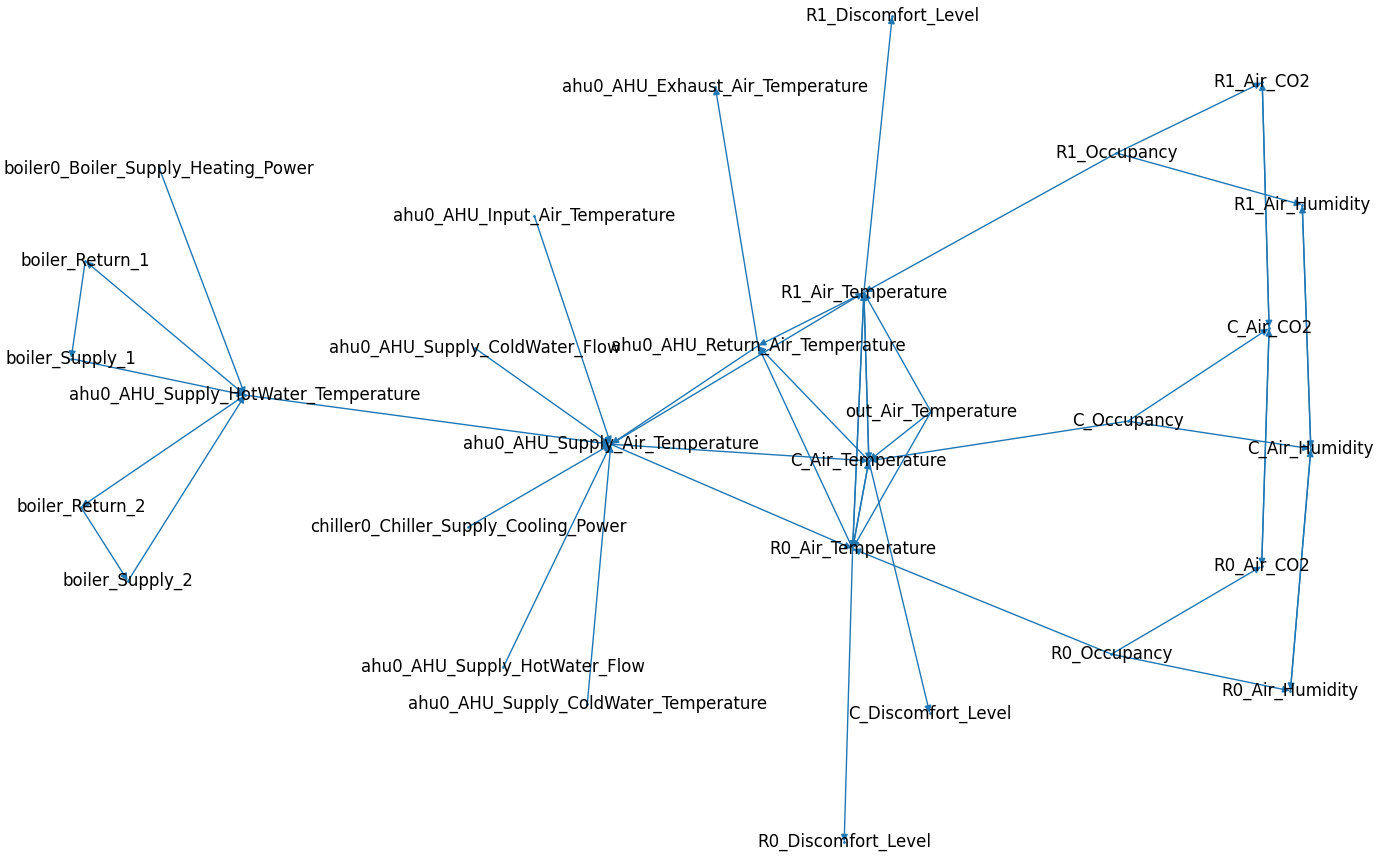}}
\vskip -0.15in
\caption{Directed graph representation of the ground truth adjacency matrix based on the physics of the simulation \cite{ploennigs2017semantic}}
\label{fig:graph-ground-truth}
\end{center}
\vskip -0.2in
\end{figure*}

\begin{figure*}[hb]
\begin{center}
\centerline{\includegraphics[width=\textwidth]{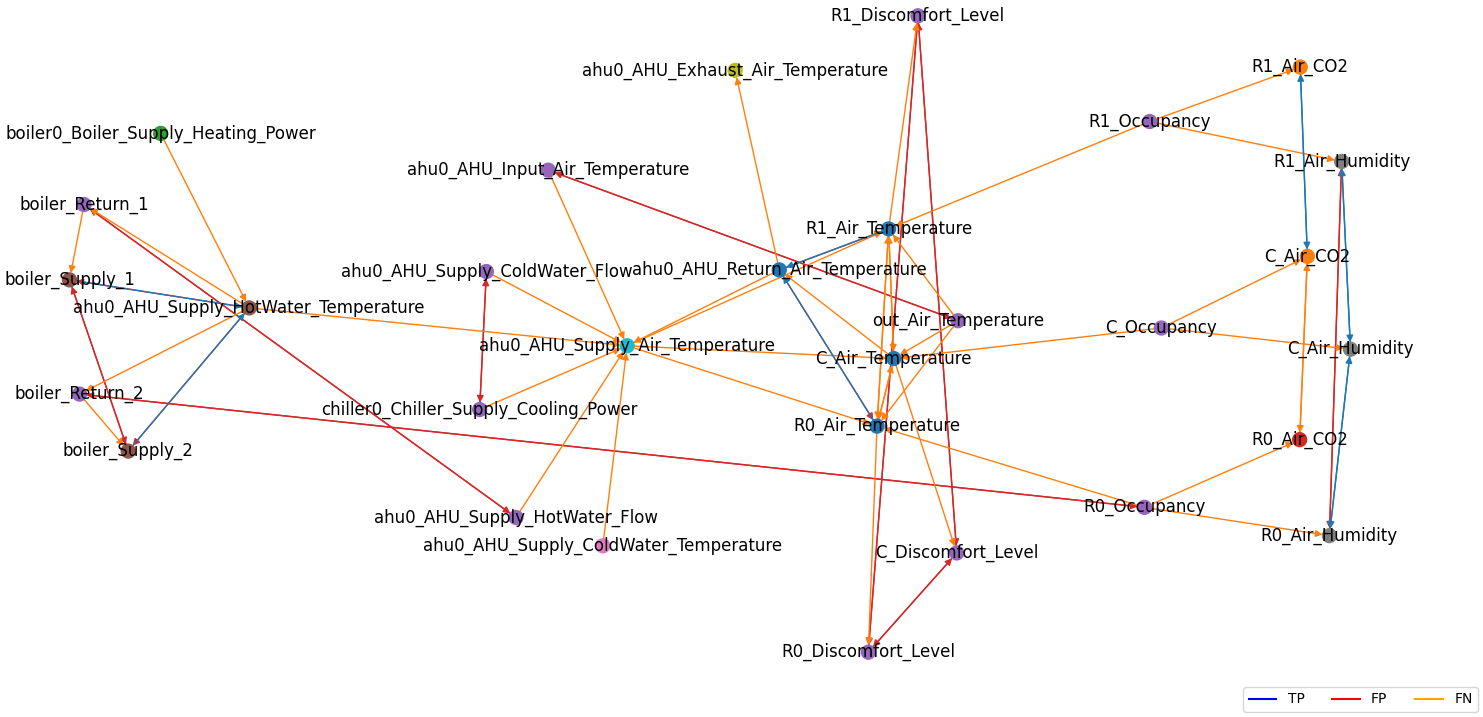}}
\vskip -0.2in
\caption{Directed graph representation of the adjacency matrix for one day using the spatial-temporal approach showing correctly identified causal relationships (TP), missing (FN), or incorrectly identified ones (FP)}
\label{fig:graph-spationtempD1}
\end{center}
\vskip -0.3in
\end{figure*}

Figures~\ref{fig:graph-ground-truth} and \ref{fig:graph-spationtempD1} show a graph representation of the ground truth adjacency matrix and the adjacency matrix discovered for a single day with the spatial-temporal approach for the building dataset.

The ground truth Figures~\ref{fig:graph-ground-truth} shows the time series of the dataset as nodes and the correct causal relationships as directed edges. Its has on the left the boiler system that is feeding a HVAC in the center which is heating/cooling the rooms on the right with occupants. 

The goal of the causal discovery is to identify this graph from the time series data. We split all time series into days and compute for each day an adjacency matrix by clustering the timeseries with DTW in smaller clusters, within which we compute the causal relationships that form the adjacency matrix.

Figure~\ref{fig:graph-spationtempD1} shows the adjacency matrix resulting from the first day as example. The edges here color-code if the edge is also in the ground truth graph (TP), not in it (FP), or missing (FN). The node color shows the clusters the time series belongs to. We can identify clusters around similar time series semantics and characteristics like CO2 cluster, a humidity cluster, or a temperature cluster. Most of the missing edges (FN) are between clusters as we ignore them for performance reasons. The additionally detected ones (FP) are within a cluster and occur in some cases where there is actually a multi-hop relationship passing trough another cluster. Good example here are the incorrect relationships around the AHU\_Supply\_Air\_Temperature. It is isolated in its own cluster without causal relationships to other nodes. To cope with this, the causal discovery identifies additional relationships that bypass this isolated node. These bypass relationships are in this case not necessarily wrong and thus also do not have negative effects on the prediction performance of the TGCN.

\end{document}